\documentclass[journal]{IEEEtran}

\usepackage{graphicx}
\usepackage{amsmath}

\begin{document}

\title{\LARGE{doScenes: An Autonomous Driving Dataset with Natural Language Instruction for Human Interaction and Vision-Language Navigation}}

\author{Parthib Roy, Srinivasa Perisetla, Shashank Shriram, Harsha Krishnaswamy,  Aryan Keskar, Ross~Greer
\thanks{Corresponding author R. Greer is with the Department of Computer Science and Engineering, University of California, Merced. e-mail: rossgreer@ucmerced.edu. All authors are members of the Machine Intelligence, Interaction, and Imagination Lab ($\text{Mi}^3$) at the University of California, Merced.} 
}

\maketitle

\begin{abstract}
Human-interactive robotic systems, particularly autonomous vehicles (AVs), must effectively integrate human instructions into their motion planning. This paper introduces doScenes, a novel dataset designed to facilitate research on human-vehicle instruction interactions, focusing on short-term directives that directly influence vehicle motion. By annotating multimodal sensor data with natural language instructions and referentiality tags, doScenes bridges the gap between instruction and driving response, enabling context-aware and adaptive planning. Unlike existing datasets that focus on ranking or scene-level reasoning, doScenes emphasizes actionable directives tied to static and dynamic scene objects. This framework addresses limitations in prior research, such as reliance on simulated data or predefined action sets, by supporting nuanced and flexible responses in real-world scenarios. This work lays the foundation for developing learning strategies that seamlessly integrate human instructions into autonomous systems, advancing safe and effective human-vehicle collaboration. We make our data publicly available at https://www.github.com/rossgreer/doScenes
\end{abstract}


\IEEEpeerreviewmaketitle

\section{Introduction}

\IEEEPARstart{T}{here} is a growing need for robotic systems, especially autonomous vehicles, to be human-interactive. In this research, we particularly focus on \textit{human-vehicle instruction interactions}, where a human agent communicates a directive to a vehicle that should influence the vehicle's motion plan. While many of the principles discussed in this research extend more generally to human-robot instruction interactions; we focus on autonomous vehicles as a special case of robot whose motion plans exists in a particular scale of time and velocity, necessitating but also benefiting from domain-specific characterizations of instructions. 

Existing interactions of humans and vehicles can be characterized by a set of attributes such as source position \cite{trivedi2007looking}, modality, referentiality, and temporality. Example options within these attributes are summarized in Table \ref{classes}.

Instructions may be described by combinations of these attributes, and options within an attribute are not always mutually exclusive and may be integrated in various combinations. For example: 
\begin{itemize}
    \item A passenger may point to a curb cut and ask to be dropped off there, using verbal and gesture-based interaction from inside the vehicle, and providing a short-term instruction which refers to a static object in the scene. 
    \item A firefighter may ask a vehicle to move out of the way, using verbal instruction from outside the vehicle, and providing a short-term instruction which refers to dynamic scene objects. 
    \item A police officer may use their whistle and hand-gestures to get a driver's attention and wave their vehicle through while directing traffic, using a combination of pseudo-verbal and gesture-based interaction from outside the vehicle, and providing a short-term instruction which does not refer to additional objects. 
\end{itemize}

\begin{figure}
    \centering
    \includegraphics[width=.5\textwidth]{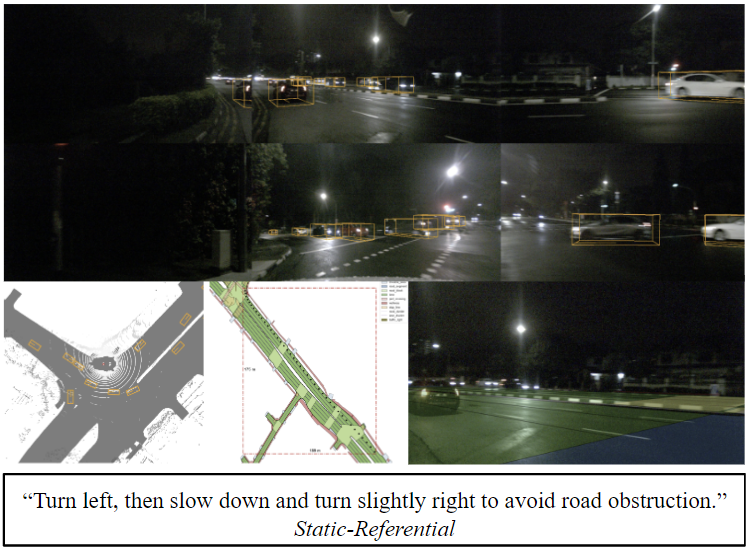}

    \caption{Typical nuScenes data includes 3D bounding box annotations, LiDAR point clouds, and driving area map feature layers. In the doScenes dataset, we augment each clip of temporal data with an instruction and a tag to indicate the instruction's referentiality.}
    \label{fig:example}
\end{figure}

In these examples, and in this research, we focus on short-term interactions, which anecdotally apply on the order of less than 10 seconds of motion. More specifically, these types of instructions contain all relevant information within a viewable proximity (e.g. no relevant landmarks or information beyond the horizon of the driver's egocentric view). The time itself is not a strictly-defined boundary. For example, in the nuScenes dataset, samples have 12 seconds of motion; sometimes, these 12 seconds stay within a visible horizon from the temporal origin, and other times, the vehicle effectively moves to an entirely new scene within 12 seconds\footnote{This point is discussed further in the section of the paper \textit{Considerations for Application and Evaluation}.}.  
Towards the development of new learning strategies in autonomous perception and planning, we introduce the doScenes dataset, a novel dataset which pairs sensor feeds, vehicle trajectories, and map information with human-interactive instructions and referentiality tags. doScenes draws its name from Judea Pearl's \textit{do-calculus} \cite{pearl1995causal}, developed to identify causal effects; accordingly, the human instructions provided in this dataset are intended to emulate commands that would cause the resulting sequence of actions taken by the driver.

\begin{table}[]
    \caption{Attributes of Human-Vehicle Instruction Interactions}
    \centering
    \begin{tabular}{|c|c|}
        \hline
\textbf{        Source Position} & Inside Vehicle, Outside Vehicle \\ \hline
        \textbf{Modality} & Voice, Gesture \\  \hline
        \textbf{Referentiality} & None, Static Objects, Dynamic Objects \\  \hline
        \textbf{Temporality} & Short-Term, Medium-Term, Long-Term \\ \hline
    \end{tabular}
    \label{classes}
\end{table}

\section{Related Research}

\subsection{Datasets for Human-Robot Instruction}


Before discussing datasets specific to autonomous driving, we first present related research in the more general field of Human-Robot Interaction (HRI), and specifically instruction-style interaction for real-world scenarios. One such dataset is NatSGD \cite{natSGD}, a multimodal dataset designed to emulate natural human communications through speech and gestures. NatSGD is primarily designed to enable robots to understand and execute real-world tasks in a natural manner, including those requiring nuanced household robotics actions like cooking and cleaning. It stands out as one of the first datasets to encompass speech, gestures, and demonstration trajectories. In the NatSGD framework, robot behaviors were developed by creating a photorealistic simulated environment using Unity3D in conjunction with a customized Robot Operating System (ROS) plugin. Additionally, real-time inverse kinematics for the robot's head movement and arms were implemented using BioIK. During tasks, the robot maintained eye contact with the target and returned its gaze to the participant when ready for the next interaction.

Another high-volume and diverse dataset is BridgeData V2 \cite{bridgeDataV2}. BridgeData V2 distinguishes itself by offering coverage across numerous tasks and domains in robotic learning research, including support for task conditioning through goal images or natural language instructions. The dataset includes over 60,000 trajectories (50,365 expert demonstrations and 9,731 from a randomized scripted policy) collected across 24 environments. Data collection used an accessible, low-cost robot platform, making BridgeData V2 appealing for academic research.  The robot setup consisted of a WidowX 250 robot arm (fixed-based) and numerous cameras, including one RGBD camera for sensing, two RGB cameras with randomized poses during data collection, and one RGB camera attached to the robot's wrist. In addition, the robot was controlled via a VR controller. Data collection occurred in indoor environments, with the majority being in toy kitchens. However, despite its scale and variety of tasks, the dataset is limited to low-precision and non-real-time activities, quite different from the requirements of autonomous driving.

The HandMeThat \cite{handMeThat} benchmark assesses instruction understanding and task execution within physical and social contexts, emphasizing situations and instructions with ambiguity. Each episode of the text-based dataset contains a sequence of steps taken by the human, followed by an instruction. The authors propose two stages for modeling robot responses. In the first stage, a robot agent observes a human agent and its actions and attempts to infer their end goal; in the second stage, the human provides a language-based instruction to the robot, and the robot acts within the environment to complete its tasks. The robot agent needs to consider both the human’s historical actions as well as the subgoal specific to human utterance. It is important to note that the benchmark lies within its operation of a text-only environment, strongly limiting its scope for vision-based environments, and does not address non-verbal communication or dynamic interactions.







\subsection{nuScenes and Natural Language in Autonomous Driving Datasets}

nuScenes \cite{caesar2020nuscenesmultimodaldatasetautonomous} is a multimodal dataset designed to fill the gap of capturing diverse real-world conditions necessary for building robust autonomous driving perception systems. At the time of its release, nuScenes was the largest AV dataset to feature a complete 360° field of view (FOV) AV sensor suite, including 6 cameras, 5 radars, and 1 lidar, and is also the first to include radar data using an AV approved for public roads. Additionally, nuScenes was the first multimodal dataset to capture nighttime and rainy condition data. The dataset includes 1,000 manually selected scenes from two highly challenging and dense traffic environments: Boston (Seaport and South Boston) and Singapore (One North, Holland Village, and Queenstown). Each scene is annotated at 2 Hz, resulting in 1.4 million 3D bounding boxes for 23 object classes. The AV utilized during data collection were two Renault Zoe supermini electric cars, equipped with front and side cameras with a 70° FOV, offset by 55°, and a rear camera with a 110° FOV.

Though novel in its instruction and interactivity basis, doScenes is not the first dataset which features nuScenes annotations extended using natural language. nuScenes-QA \cite{qian2024nuscenes} combined nuScenes' 3D detection annotations with question templates, automatically generating 460K question-answer pairs based on scene graphs. nuScenes-MQA (Markup Question Answering) \cite{inoue2024nuscenes} introduced questions and answers enclosed within markups of particular objects within the visual scene. 



Beyond re-annotations of nuScenes, other datasets have been developed to integrate natural language information into the driving environment. The Rank2Tell dataset \cite{sachdeva2024rank2tell} advances autonomous driving with multimodal data annotated with visual elements in a traffic scene ranked by relevance to safety, traffic rule compliance, and the dynamic context of the situation. By emphasizing contextual prioritization, Rank2Tell provides a benchmark for evaluating how well autonomous vehicle systems align with human judgment. While Rank2Tell makes significant contributions to ranking-based reasoning and highlights the importance of competing visual elements, it is limited to scene-level understanding and is only based on certain specific key-frames of the traffic scene video. It lacks actionable instructions for motion planning, restricting its applicability to real-world driving scenarios where vehicles must respond to specific directives. doScenes addresses this gap by bridging the divide between multimodal reasoning and actionable instructions. Annotations can be directly tied to objects of importance in traffic scenes, focusing on the execution of human and natural-language commands. By complementing Rank2Tell’s emphasis on importance ranking with actionable directives, doScenes offers a novel framework for training and evaluating autonomous systems in real-world human-vehicle interactions.

The GPT-Driver framework \cite{mao2023gpt} transforms autonomous driving motion planning into a language modeling task using OpenAI’s GPT-3.5 model. It converts inputs like sensor data and vehicle states into a unified language representation, generating driving trajectories as language tokens with natural language explanations. This tokenized-driving approach enhances interpretability and generalization, allowing for greater transparency in decision-making. Tested on the nuScenes dataset, GPT-Driver achieved state-of-the-art trajectory prediction accuracy on the nuScenes dataset with a centimeter-level L2 error and competitive collision rates. However, it is evaluated in an open loop form on 3-second intervals, and does not make use of important features found in doScenes, such as egocentric views or human instruction beyond generic high-level objectives (e.g. ``right"), limiting its ability to generalize to novel scenarios or instructions.

With similar data to doScenes, the DriveMLM and LMDrive frameworks \cite{wang2023drivemlm, shao2024lmdrive} leverage large language models (LLMs) for autonomous driving by integrating multimodal data from CARLA simulations, including images, LiDAR data, traffic rules, and user commands, to align with human instructions. The primary aim is to enable the LLM to predict actionable steps for the ego vehicle to execute given specific human commands, a similar objective to doScenes. DriveMLM is supported by a robust dataset of 280 hours of annotated CARLA simulation data, including decision states and corresponding explanations. LMDrive enables vehicles to perform step-by-step navigation tasks, supported by a 64K-clip dataset encompassing diverse scenarios and complex instructions. LMDrive segments training data into clips, with each clip corresponding to one navigation instruction from a pre-defined set of 56 instructions. Notably, they augment their dataset by using ChatGPT to generate 8 semantically-equivalent variants of each instruction. 

However, the reliance of DriveMLM and LMDrive on simulated data restricts their applicability to real-world scenarios. Additionally, the frameworks prioritize immediate, one-step decisions over multi-step planning or adaptive driving styles necessary for handling complex environments, reflected in the limited output of DriveMLM within a set of instructions such as \textit{keep, accelerate, left change, right change}, which may fail to fully capture the nuances and complexities of diverse driving scenarios and constrain the system’s flexibility in responding to unique situations requiring dynamic or nuanced decision-making. doScenes does not limit instructions to a fixed set; annotators can freely describe each scene's instructions, and because there are multiple ways to give an instruction with the same effect, some choose to make references to scene objects while others do not. doScenes brings the concepts of DriveMLM and LMDrive to real-world data and provides instructions which serve decisions beyond one-step decisions, further augmented with tagged information on whether instructions rely on dynamic or static objects within the driving scenes. We note that, at the time of writing, only LMDrive has made their data publicly available.

DriveGPT4 \cite{xu2024drivegpt4} is an interpretable end-to-end autonomous driving system using LLMs for multimodal reasoning and control. It integrates video inputs and textual queries to predict low-level vehicle control signals and provides natural language explanations for its actions. Motivating our research, the authors note the scarcity of publicly available datasets suitable for their task, and train on an enhanced BDD-X dataset with ChatGPT-generated Q\&A pairs for questions equivalent to \textit{what} is the vehicle doing, \textit{why} is the vehicle doing it, and \textit{what will} the vehicle do next. DriveGPT4 excels in action description, justification, and control signal prediction, offering actionable decisions and user-friendly explanations. However, the dataset is again without human instruction, only explanation of the current state. 

The DRAMA dataset \cite{DRAMA} advances situational awareness in autonomous driving by addressing two key challenges: identifying risks and explaining them. DRAMA (Driving Risk Assessment Mechanism with A captioning module) distinguishes itself by not only pinpointing risks in driving scenes but also describing them in natural language. It contains 17,785 real-world driving scenarios from urban roads in Tokyo, Japan, with detailed annotations on risks, critical objects, and interactions from the driver’s perspective. By combining visual reasoning with linguistic explanations, DRAMA lays a foundation for enhancing perception and communication systems in autonomous vehicles. DRAMA excels in localizing and explaining risks, associating important objects like vehicles or pedestrians with potential hazards and pairing these insights with natural language captions. It benchmarks multi-task models that simultaneously identify risks and generate explanations, fostering advancements in vision-language integration. However, DRAMA's natural language annotations focus on assessing and understanding risks, whereas doScenes is built around driving instructions. 




\subsection{Bridging Human-Robot Instruction and Autonomous Driving}

Unlike previously-mentioned datasets which emphasize scene understanding and description, our research is the first public real-world dataset to provide \textit{driving instructions and referentiality information} as the natural language annotation, creating a link between imperative language and motion for autonomous vehicles. This task is relevant for autonomous vehicles due to sudden changes to the environment which can create novel or anomalous scenarios, even in spatial locations which may have been typical in moments prior. Vision-language models have been successful in detecting such scene changes \cite{greer2024towards}. Such anomalies may require a manual takeover response \cite{rangesh2021autonomous, greer2023safe, rangesh2021predicting}, but in cases where a driver is unable to operate the vehicle, the ability of the vehicle to autonomously respond to commands can be especially valuable; this task of navigation of a novel environment without a map but with natural language instruction is often referred to as Vision-and-Language Navigation (VLN) \cite{cheng2024navila, anderson2018vision}. The task of translating language to actuated action is complex, requiring reasoning, closed-loop planning, and control. NaVILA \cite{cheng2024navila} addresses this task by adding an intermediate mid-level, language-based representation of actions above low-level control signals, providing a reasonable intermediate in the process of translating language to action. This decoupling of language and control allows one VLA to modularly fit a new robot with appropriate adjustment of low-level control policy, particularly useful towards an autonomous driving setting where different vehicles may have different control configurations. NaviLLM \cite{zheng2024towards} frames the task as a VQA, with an answer set comprised of scene views from multiple directions; the predictive task is to identify which direction the robot should move next toward the completing of a text-prompted goal, evaluating based on whether (and how soon) the robot reaches the desired goal. Essentially, this evaluation is a step-through of panoramic options, which is infeasible for open-world driving where additional agents affect safe reachability of spatial states. With similar one-to-one framing of waypoints as images observed by the robot, LM-Nav \cite{shah2023lm} uses a goal-conditioned model to infer a sequence of graph vertices to traverse for the purpose of visiting landmarks identified by an LLM and grounded to the scene \cite{xu2021grounding} by a VLM. This approach requires that the robot first collects image and GPS observations over the intended search area. This is, however, less practical for a dynamic environment where landmarks may be other agents themselves, and have no grounding to a map, something reflected in the \textit{dynamic-referential} instruction subset of doScenes. The algorithm of Hu et al. \cite{hu2019safe} may better interact with such dynamism by grounding constraint-based instructions through detection-informed adjustments to a costmap which is then used for collision-avoidant navigation. As in additional works, goals are grounded to locations in a predefined semantic map for global motion plans.

\section{doScenes Dataset}

The process of collecting and annotating driving-instruction data is a complex task; a test vehicle must be properly equipped with appropriate sensors for perception, and interior microphones must capture and synchronize verbal commands to driving events. Following collection, expensive annotation of objects and visual scene features must occur to enable supervised learning. Fortunately, massive datasets such as nuScenes provide completion of large portion of this task (that is, large-scale collection and vision-based annotation). We apply retroactive annotation of driving instructions by playing back each of the 1,000 12-second nuScenes clips, and transcribing an instruction (or lack of instruction) that would be given to a driver from the vantage of the passenger to initiate the motion plan observed in the clip. 

This natural language instruction can be generated by a heuristic we name the \textit{taxi test}: if you were being driven through this scene by a taxi driver, what instruction, if any, would you need to give an instruction to trigger the behavior observed in the video? 

For each of the 1,000 scenes in nuScenes, we provide a set of instruction annotations. These annotations are generated by five independent annotators, and each annotator may include multiple annotations for a scene if they imagine multiple instructions which may generate a similar series of events. 

An instruction field may be blank if no instruction interaction is needed to `cause' the action (e.g. waiting at a red light, continuing in your lane with the flow of traffic, etc.). Referring to the taxi test, instructions should instigate a change from default vehicle motion. 

In addition to the annotated instruction, we provide an additional column for instruction referentiality. When an instruction refers to dynamic objects, e.g. ``follow the white van", the \textit{dynamic reference} tag is given. When an instruction refers to static objects, e.g. ``stop at the blue sign", the \textit{static reference} tag is given. It is possible for an instruction to be annotated with zero, one, or both of these tags. 

We note that even though an instruction may not be referential, it is still expected that the autonomous vehicle (or driver) is fully aware of the major static and dynamic objects in the scene at all times. This is a prerequisite for safe autonomous driving, independent of instruction interactions. Rather, the dynamic referentiality tag is intended to indicate which instructions may require further observation of an object than is available at the moment of instruction. Such instructions cannot be evaluated in an ``open loop" manner, since the playout of the scene's dynamic objects will influence the ego motion plan.

\begin{figure}
    \centering
    \includegraphics[width=0.45\textwidth]{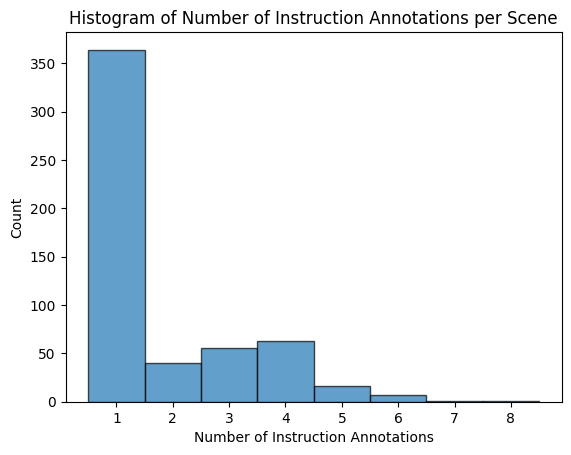}
    \caption{Histogram of number of instruction annotations per scene; most of the scenes of doScenes have only one or two annotations. Having a greater number of instruction annotations reflects an annotator's generation of multiple possible instructions that could cause the same scene playout.}
    \label{fig:hist}
\end{figure}

\section{Considerations for Application and Evaluation}

\begin{table}[]
    \centering
        \caption{Statistics on Referentiality of doScenes Instructions}
    \begin{tabular}{c|c}
        Non-Referential & 535 \\
        Static Referential & 214 \\
        Dynamic Referential & 159 \\
        Both & 93 \\
    \end{tabular}

    \label{tab:my_label}
\end{table}

In this section, we provide some observations on the nature of the annotated instructions and their relationship to nuScenes data, with possible implications in how this data may be useful for learning, and where some limitations may exist.

While the instructions in doScenes provide information about where a vehicle should move, it does not necessarily provide information about how the vehicle should move, e.g., driving speed or style. This is primarily due to the retroactive annotation approach applied; in future datasets, instructions on driving style or speed may be provided to instigate particular driving responses. Accordingly, motion plans generated by natural language learned from doScenes may not be responsive to prompts related to speed or style. 

Further, the 12-second duration of the scenes in the nuScenes dataset, especially when taken at free-flowing urban traffic speeds, may present motion plans longer than a single instruction can cover. This should be accounted for when using doScenes as a basis for evaluation; accurate response to a prompt may be reflected in only the first $t$ seconds of a nuScenes path before the instruction becomes irrelevant after a significant change of scenery or transition to later stages of a multi-step motion plan. This also opens for future research the consideration of frameworks for multi-stage motion planning using natural language. 

We chose to create the tags for static and dynamic referential instructions so that motion plans and associated models can be trained or evaluated over particular sets (e.g. those without reference to certain types of objects). For example, models which use only the rasterized map as input, which have found decent success in prior trajectory prediction tasks \cite{deo2022multimodal, gilles2022thomas, greer2021trajectory}, may be able to learn appropriate motion plans for non-referential instructions, but would be limited without the LiDAR or front-view image as input for making sense of object references.

doScenes was designed to provide data for systems to learn a relationship between instructions and vehicle motion. If such a model can be learned, future research may include techniques to use this model to generate a trajectory based on natural language, or assign a natural language descriptor to a vehicle trajectory, contributing to the task of interpretable, interactive autonomous vehicle motion planning. 
 
As an example of models which may be extended from vision-language to vision-language-action based on prompted instruction, SpatialRGPT \cite{cheng2024spatialrgpt} learns representations at the instance level (rather than global level) from 3D scene graphs and integrates depth information to enhance VLMs’ spatial perception and reasoning capabilities. Importantly, nuScenes provides 3D input in the form of LiDAR point clouds, making it an appropriate dataset for this VLM, and the annotations of doScenes create possibilities for learning information about the corresponding actions to instructional inputs, which can be explored as future research enabled by this dataset, crossing from the generalized robotics domain to the specific challenges of autonomous driving. 

\section{Concluding Remarks and Future Research}

In addition to the application areas for future research identified throughout the paper, in this section, we would like to highlight future research potential for data collection beyond doScenes. doScenes is a novel form of autonomous driving data where natural language instructions are paired with driving scenes and respective sensor time series. However, the instructions in this case are annotated retroactively; while the annotators give their best estimate of an instruction that would have caused such a scene to unfold, this is only a proxy for a true signal. Future data collection should pair true human instructions with action responses. There are a variety of settings (both naturalistic and experimental) which can allow for such collection, and it is reasonable to expect that a higher volume of such high-quality data will enable better learning of corresponding VLA models.

\section*{Acknowledgment}

The authors thank $\text{Mi}^3$ lab members for their contribution to the annotations of the dataset. 

\bibliographystyle{ieeetr}
\bibliography{refs}







\end{document}